\documentclass[lettersize,journal]{IEEEtran}
\usepackage{amsmath,amsfonts}
\usepackage{algorithmic}
\usepackage{algorithm}
\usepackage{array}
\usepackage{textcomp}
\usepackage{stfloats}
\usepackage{url}
\usepackage{verbatim}
\usepackage{graphicx}
\usepackage{cite}
\usepackage{hyperref}
\usepackage{microtype}
\usepackage{subfigure}
\usepackage{booktabs} 
\usepackage{amssymb}
\usepackage{multirow}
\usepackage{ulem}
\usepackage{fancyhdr}
\usepackage{amssymb}
\usepackage{mathtools}
\usepackage{paralist}
\usepackage{siunitx}
\usepackage{soul}
\usepackage{url}
\usepackage{textcomp}
\usepackage{verbatim}

\usepackage[makeroom]{cancel}

\newcommand{\calvin}[1]{{\textcolor[rgb]{0.7,0.3,0.0}{$^{\text{CM}}$: {\it #1}}}}

\newcommand{\bfloat}{{\sc bfloat16}}
\newcommand{\floatfp}{{\sc float32}}
\newcommand{\floathp}{{\sc float16}}
\newcommand{\intqp}{{\sc int8}}

\DeclarePairedDelimiter{\ceil}{\lceil}{\rceil}

\DeclarePairedDelimiter{\round}{\lfloor}{\rceil}
\DeclarePairedDelimiter\abs{\lvert}{\rvert}

\newcommand{\beginsupplement}{%
        \setcounter{table}{0}
        \renewcommand{\thetable}{S\arabic{table}}%
        \setcounter{figure}{0}
        \renewcommand{\thefigure}{S\arabic{figure}}%
}

\begin{document}

\title{Adaptive Block Floating-Point for Analog Deep Learning Hardware

\author{
\IEEEauthorblockN{
Ayon Basumallik,
Darius Bunandar,
Nicholas Dronen,
Nicholas C. Harris,
Ludmila Levkova,
Calvin McCarter,
Lakshmi Nair,
David Walter,
David Widemann
}\thanks{All the authors are affiliated with Lightmatter, Inc., Boston, Massachusetts, USA}
}

}
\markboth{Journal of \LaTeX\ Class Files,~Vol.~14, No.~8, August~2021}%
{Shell \MakeLowercase{\textit{et al.}}: A Sample Article Using IEEEtran.cls for IEEE Journals}

\IEEEpubid{0000--0000/00\$00.00~\copyright~2021 IEEE}

\maketitle
\thispagestyle{plain}
\pagestyle{plain}

\begin{abstract}
Analog mixed-signal (AMS) devices promise faster, more energy-efficient deep neural network (DNN) inference than their digital counterparts. However, recent studies show that DNNs on AMS devices with fixed-point numbers can incur an accuracy penalty because of precision loss. To mitigate this penalty, we present a novel AMS-compatible adaptive block floating-point (ABFP) number representation. We also introduce amplification (or gain) as a method for increasing the accuracy of the number representation without increasing the bit precision of the output. We evaluate the effectiveness of ABFP on the DNNs in the MLPerf\texttrademark{} datacenter inference benchmark---realizing less than $1\%$ loss in accuracy compared to \floatfp{}. We also propose a novel method of finetuning for AMS devices, Differential Noise Finetuning (DNF), which samples device noise to speed up finetuning compared to conventional Quantization-Aware Training.
\end{abstract}


\section{Introduction}
\label{intro}

The power consumption and carbon footprint of datacenters used to run compute-intensive deep neural networks (DNNs) have grown substantially in the last decade. Most of the future datacenter workload is expected to come from DNN inference, which comprises 80--90\% of the total compute time consumed during a DNN's lifespan, from training to deployment~\cite{patterson2021carbon, dlinference_fb}. The growing demand of DNN inference is associated with a widening range of applications and increasingly larger DNNs, which require a larger amount of compute and memory.

In light of the above, computing technologies with potential advantages in speed and energy efficiency, such as analog mixed-signal (AMS), are worth exploring. AMS performs compute-intensive matrix multiplications in the analog domain while running other operations---notably, non-linearities and data storage---in the digital domain. Matrix multiplication is the lion's share of a DNN's inference operations~\cite{dally2015high} because of its computational complexity: approximately $O(n^3)$ for inputs of size $n \times n$ elements. By encoding information in the analog domain---in the amplitude, phase, or time of a physical signal---AMS devices have the potential to use less energy per compute operation than their digital-only counterparts, albeit at the cost of a lower bit precision and possible susceptibility to noise. This can be a challenge for applications requiring noiseless or high-precision compute, such as cryptographic hashing. DNNs, however, tend to be more robust to noise, making them generally amenable to running on AMS devices~\cite{seok2019cases}.

Prior works in running DNNs on AMS devices have focused on convolutional neural networks (CNNs). Previous results show that low-precision ($<8$ bit) activations and weights are sufficient for small CNNs, such as for classifying MNIST and CIFAR-10 images \cite{deng2020model}. Recent works have started to explore the use of AMS for larger CNNs, e.g., ResNet-50 for ImageNet classification~\cite{rekhi-dally-2019}, and small language models, e.g., recurrent neural networks (RNNs) trained on Penn TreeBank ~\cite{ghodrati2020}.

\begin{table}[t]
    \vspace{-0.1in}
    \centering
    \caption{MLPerf\texttrademark{} datacenter inference benchmark.}
    \vspace{0.1in}
    \begin{tabular}{lll}
    \bf Task                 & \bf DNN         & \bf Dataset     \\
    \hline
    Image classification & ResNet50     & ImageNet    \\
    Object detection     & SSD-ResNet34 & MS COCO     \\
    Image segmentation   & 3D U-Net     & BRaTS 2019  \\
    Speech recognition   & RNN-T        & Librispeech \\
    Question answering   & BERT Large   & SQuADv1.1 \\
    Recommendation       & DLRM         & 1TB Click Logs      \\
    \end{tabular}
    \label{tab:ml_commons_tasks}
    \vspace{-0.2in}
\end{table}

The results showed that relatively large-precision activations (with $\geq 11$ bits) are required to achieve quality commensurate ($<1\%$ loss) to that achieved with \floatfp{}~\cite{rekhi-dally-2019}. Working with an analog-to-digital converter (ADC) that operates with $\geq 11$ bits of precision at gigahertz rates can be prohibitive as the power consumed by the mixed-signal converters scale exponentially with the bit precision ($\sim 2^b$, where $b$ is the number of output bits). Alternatively, the grade-school multiplication technique of partitioning each matrix multiplication into several multiplications with lower bit-precision operands can reduce the required bit precisions \cite{ghodrati2020}. However, this technique limits the size of the matrix tile that can be programmed as the ADC bit precision must be sufficient to capture the full precision of the partial multiplications. Overall, this multiplication technique may reduce the speed of the AMS device. These limitations, however, occur because previous techniques have focused only on entirely fixed-point-precision networks.

\begin{figure*}[t!]
\centering
\includegraphics[width=0.7\textwidth]{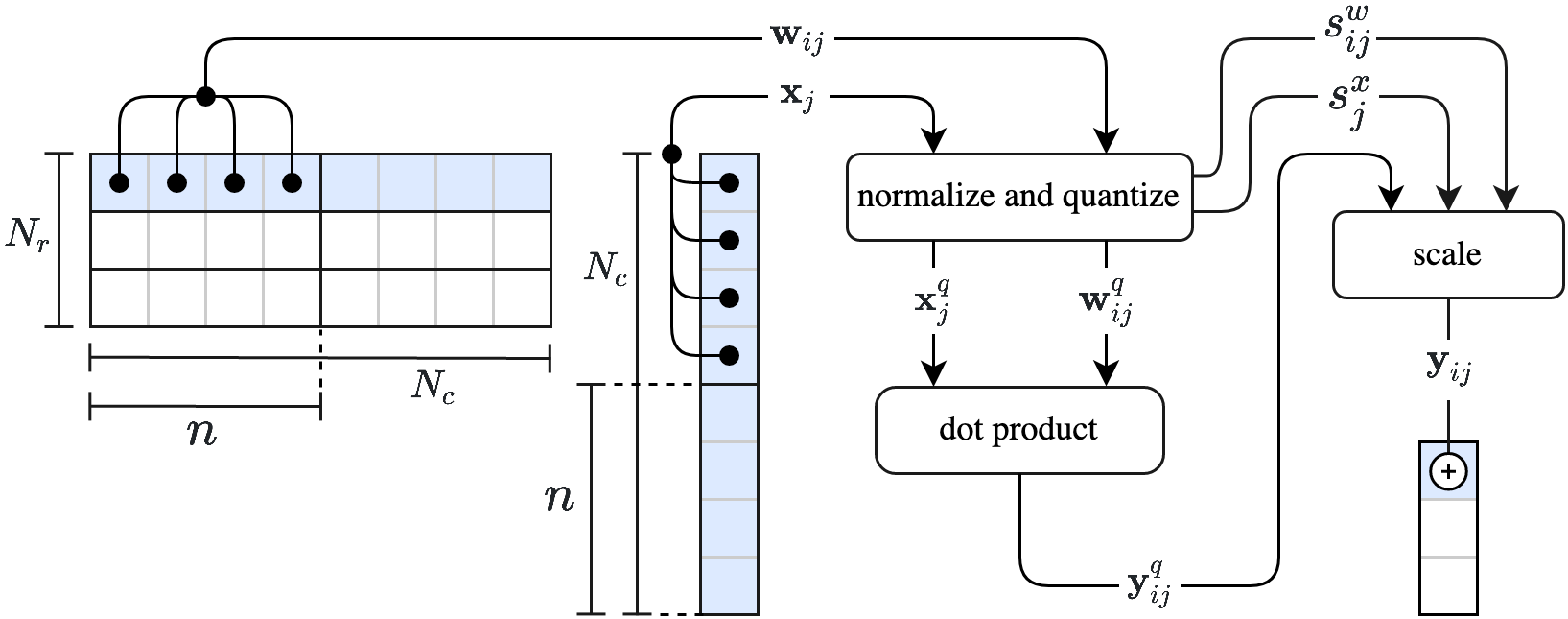}
\vspace{-0.1in}
\caption{Tiled Matrix Multiplication with Adaptive Block Floating-Point (ABFP). The row vector $\mathbf{w}_{ij}$ and column vector $\mathbf{x}_{j}$ are normalized with scales $s^{w}_{ij}$ and $s^{x}_{j}$ and quantized before the dot product operation. The output of the dot product $\mathbf{y}^{q}_{ij}$ is scaled to $\mathbf{y}_{ij}$ before it is accumulated in the final output $\mathbf{y}_{i}$. The entire pipeline, from $\mathbf{w}_{ij}$ and $\mathbf{x}_{j}$ to $\mathbf{y}_{ij}$, may be understood as computing a partial dot product, the parts of which are accumulated into the final output $\mathbf{y}_{i}$.}
\label{fig:tiled_gemm_figure}
\end{figure*}

\IEEEpubidadjcol
The purpose of this paper is to demonstrate an AMS design that achieves high network-level quality\footnote{``Quality'', in this paper, refers to a task-specific metric, such as accuracy or F1 score.} without the need for high bit-precision analog/digital converters or grade-school multiplication schemes. At the core of our approach is the novel adaptive block floating-point (ABFP) number representation that reduces quantization effects by scaling the vectors of length $n$ within the inputs to a tiled matrix multiplication. We evaluate our approach on six unique, industry-relevant tasks, DNNs, and datasets, as shown in Table~\ref{tab:ml_commons_tasks}.

The main contributions of this paper are:
\begin{enumerate}
    \item a novel adaptive block floating-point (ABFP) number representation;
    \item the use of amplification (or gain) as a method for increasing the accuracy of the number representation without increasing the bit precision of the output; and
    \item two viable methods for finetuning DNNs---the conventional Quantization-Aware Training (QAT) and the novel Differential Noise Finetuning (DNF)---to achieve high quality despite quantization error and noise.
\end{enumerate}

The rest of the paper is organized as follows. Section \ref{related} discusses related work. Section \ref{sec:abfp} details the ABFP based approach. Section \ref{sec:finetuning} details approaches to finetuning using QAT and DNF. Section \ref{results} presents a quantitative evaluation of our approach using the MLPerf\texttrademark{} datacenter inference benchmarks~\cite{reddi2020mlperf, visionmlperf2021}. Section \ref{discussion} discusses the applicability and ramifications of this approach at a system level. Section \ref{conclusion} concludes the paper.

Our results show that ABFP enables DNNs to run with high accuracy on an AMS device. The six MLPerf\texttrademark{} datacenter inference benchmark DNNs lose~$<1\%$ of their \floatfp{} quality with a small tile size of $n=8$ and no additional gain. Also, prior to finetuning, four of the six DNNs achieve~$<1\%$ degradation with a large tile size of $n=128$ and appropriate gain. The other two DNNs can achieve the \floatfp{} quality after finetuning with either QAT or DNF.

\section{Related Work}
\label{related}

Approaches to making DNN inference faster and more energy efficient generally focus on reducing the cost of matrix multiplication. In software, distillation~\cite{bucilua2006model,hinton2015distilling}, pruning~\cite{lecun1990optimal, blalock2020state}, and dimensionality reduction~\cite{liberty2013simple,dasgupta2010sparse} have been proposed to decrease the number of multiplications or size of the operands. Scalar quantization has been used to perform reduced-precision multiplications, using standard IEEE formats such as \floathp{} or \intqp{}, newer formats such as Google Brain Float (\bfloat{})~\cite{abadi2016tensorflow}, or combinations thereof~\cite{khudia2018open}. 

Hardware strategies involve either digital or analog designs to minimize the cost of multiplication. Digital chips such as NVIDIA tensor cores~\cite{nvidia2017v100} and application-specific integrated circuits (ASICs), such as the Google TPU~\cite{jouppi2017datacenter} or the Graphcore IPU~\cite{jia2019dissecting} exploit various forms of parallelism to tackle latency, but these hold less promise for tackling energy usage.

Because multiplication and accumulation are naturally efficient operations in the analog domain, custom analog circuits for DNN inference hold promise for reducing both latency and energy.
AMS designs are currently under development both in academia \cite{seok2019cases} and industry \cite{fick2018analog}. Recent works ~\cite{rekhi-dally-2019,gonugondla2020,murmann-survey} have sought to characterize the fundamental energy-accuracy tradeoffs with analog multiplication hardware, focusing on the limits imposed by ADC noise.


The current understanding of tradeoffs in this space was established by Rekhi et~al.~\cite{rekhi-dally-2019} who proposed a general ADC noise model for dot products. In the simulated model, matrix multiplications are decomposed into dot products, each of which is computed in analog using the fixed-point number format. This analog hardware circuit is characterized by: the number of bits used to store inputs, weights, and outputs in fixed-point (which may be different from each other); the number of terms in the dot product $n$; and the ADC noise which is additive and independent of the true output value. The authors combine this noise model (which simulates the effect ADC noise has on DNN inference accuracy) with an ADC energy model of the noise-energy tradeoff, in order to derive a model for the accuracy-energy tradeoff. The analog circuit's own energy consumption is not modeled, but this is typically dwarfed by ADC energy and can be safely ignored. Finally, they propose finetuning DNNs with their ADC simulation model to improve inference accuracy.

More recently, various approaches to improve this tradeoff have been put forward. Ghodrati et~al.~\cite{ghodrati2020} proposed to reduce the effect of ADC noise via parallel bit-interleaved analog compute units sharing a single ADC. Dynamic analog precision has been proposed by Gonugondla et~al.~\cite{gonugondla2020} to adapt to the variation in noise sensitivity across different portions of network architectures. Garg et~al.~\cite{garg2021dynamic} proposed averaging  the results of multiple matrix multiplications to reduce the effect of analog device noise. Our approach departs from the aforementioned proposals. Instead, we build on previous work on block floating-point (BFP) numeric representations and on gain (signal amplification) techniques, though with substantial differences.


BFP numeric representations, first developed for signal processing applications \cite{oppenheim1970realization}, have also found recent use for digital DNN acceleration. A BFP representation for acceleration of DNNs on field-programmable gate array (FPGA) hardware has been proposed by Song et al.~\cite{song2018computation}. BFP is also used in digital DNN training \cite{drumond2018training} with tile-specific exponents and \floatfp{} for activations and weights. Neither of these techniques use adaptive scaling to recover less significant bits, as ours does. A recent method for weight quantization ~\cite{mellempudi2017mixed} used per-block scaling factors to reduce quantization error in digital inference, but unlike ours, it did not use adaptive scaling factors computed at runtime then applied to weights, inputs, and activations. As all the BFP methods above are digital techniques, they do not use gain, which is specific to analog computation.

Some digital methods for reducing information loss due to \intqp{} quantization resemble our use of gain. Clipping values above the $k$-th percentile of a distribution ~\cite{haowu2020quantization} increases precision of smaller values by saturating the largest ones. Using one scale for large values and another for small ones is another approach~\cite{jain2019}. These approaches are orthogonal to our use of gain and could be used along with ABFP to positive effect.

In terms of recovering DNN quality, there is ongoing research on faster alternatives to standard quantization-aware training (QAT) of deep networks \cite{jacob2018quantization, krishnamoorthi2018quantizing,haowu2020quantization}, which uses the Straight-Through Estimator (STE) \cite{bengio2013estimating} to cope with the non-differentiability of quantization. Fan et~al.~\cite{fan2020training} adapted QAT to weight quantization by only quantizing a random subset of weights in each batch. Baskin et~al.~\cite{baskin2021uniq} finetuned weight-quantized networks by using additive noise as a proxy for quantization (called UNIQ). However, since they addressed weight-quantization, their noise distribution is conditioned on weight magnitudes; in contrast, we introduce additive noise at the level of layer outputs.

\section{Adaptive Block Floating-Point (ABFP) in AMS Hardware}
\label{sec:abfp}

The basic block of operation on an AMS device is the dot product of two vectors. The physical nature of the analog computation limits the range of numbers that can be represented. With the use of digital-to-analog (DAC) converters and ADCs, these numbers can be represented as fixed-point numbers. Fixed-point numbers, however, have restricted dynamic range compared to floating-point numbers, which are the typical number representation for DNN training. This limits the representational power of the DNNs that can be executed on such a device. 

The BFP number representation~\cite{song2018computation} works around this restriction by introducing shared scales (or shared exponents) which normalize the vectors to be within the unit range of $[-1.0, 1.0]$ and then converts them into fixed-point numbers before the dot product. After the dot product operation, the output is multiplied with the appropriate scale. In this section, we present ABFP, our adaptive version of the BFP numerical format. 

DNN training and inference are typically performed with \floatfp{} numbers. Recently, the \bfloat{} number representation (with an 8-bit mantissa and an 8-bit exponent) was created specifically to cover a range of values similar to the range of the former, while using only 16 bits~\cite{abadi2016tensorflow}. The conversion between fixed point numbers and \bfloat{} can be easily pipelined along with the vector dot product. Since the \bfloat{} format has an 8-bit mantissa, the conversion between integers with 8 bits or less (typically used for AMS devices) is especially advantageous. Therefore, we adopt \bfloat{} as the default precision to store the scales of the vectors. We also rescale the outputs of each dot product to \bfloat{}. When the matrix is wider than the tile width, the result is a sum over the partial \bfloat{} outputs of the tiled matrix-vector multiplications. In these cases, the final sum is accumulated in \floatfp{} and then converted to \bfloat{} precision as well.

\subsection{Tiled Matrix-Multiplication with ABFP}
\label{sec:tiled_gemm_with_abfp}

Figure~\ref{fig:tiled_gemm_figure} sketches the process of performing a tiled matrix multiplication with ABFP dot products. Consider a linear layer within a feedforward network with a $N_r \times N_c$ weight matrix $\mathbf{W}$ and an input activation $\mathbf{x}$ with $N_c$ elements. Both the weight and the input are represented in \bfloat.

In our scheme, we choose a vector of length $n$ to share a single \bfloat{} scale. For the weights, each row vector of length $n$ (referred hereunto as tile width) will be represented as ABFP, and thus for each of the $N_r$ rows, there are a total of $\ceil{N_c/n}$ such vectors. We label these vectors as $\mathbf{w}_{ij}$, where the subscript $i$ denotes the row and $j$ denotes the tile. In the ABFP representation, each vector $\mathbf{w}_{ij}$ shares a single scale $s^{w}_{ij} = \max\left( \abs{\mathbf{w}_{ij}} \right)$. The normalized weight vectors are $\hat{\mathbf{w}}_{ij} = \mathbf{w}_{ij}/s^{w}_{ij}$. Similarly, for the input activation each column vector of length $n$ is represented as an ABFP. The normalized column vectors of $\mathbf{x}$ are $\hat{\mathbf{x}}_{j} = \mathbf{x}_{j}/s^x_j$ where $s^x_j = \max\left( \abs{\mathbf{x}_{j}} \right)$, and $j$ indicates the tile.

The normalized vectors $\hat{\mathbf{w}}_{ij}$ and $\hat{\mathbf{x}}_{j}$ are represented as fixed-point numbers with bitwidths $b_W$ and $b_X$, respectively. The conversion can be performed using the quantization function (for a vector $\mathbf{v}$):
\begin{equation}
    Q(\mathbf{v}; \delta_v, \tau_v) = \text{clamp} \left( \round*{\frac{\mathbf{v}}{\delta_v}} \delta_v; \tau_v \right)
\end{equation}
where the clamp function limits all elements between $[-\tau_v,\tau_v]$, i.e., $\text{clamp}(\mathbf{v}; \tau_v)=\max(\min(\mathbf{v},+\tau_v), -\tau_v)$. Here, $\delta_v = 1/(2^{b_v -1}-1)$ is the size of the discretization bin for symmetric quantization with respect to 0 (for signed integers). The rounding operation denoted by $\round{\cdot}$ uses round-half-to-even method. The vectors $\hat{\mathbf{w}}_{ij}$ are quantized with $\delta_W$ (and $b_W$) and $\tau_W=1$, and the vectors $\hat{\mathbf{x}}_{j}$ with $\delta_X$ (and $b_X$) and $\tau_X=1$, as follows:
\begin{equation}
\begin{aligned}
    \label{eq:quantization_abbr}
    \mathbf{w}_{ij}^q &= Q(\hat{\mathbf{w}}_{ij}; \delta_W, \tau_W) \\
    \mathbf{x}_{j}^q &= Q(\hat{\mathbf{x}}_{j}; \delta_X, \tau_X)
\end{aligned}
\end{equation}
Then the output of the dot product on the AMS device is:
\begin{equation}
\begin{aligned}
    \label{eq:quantization}
    \mathbf{y}_{ij}^q &= Q\left( \mathbf{w}_{ij}^q \cdot \mathbf{x}_{j}^q; n \delta_Y, \tau_Y \right)
\end{aligned}
\end{equation}
where the output is clamped with $\tau_Y=n$ due to the summation within the dot product that results in an output between $[-n, n]$. It also increases the discretization bin by a factor of $n$. This output can be converted back to \bfloat{} by multiplying it with the appropriate scale $s^y_{ij} = s^w_{ij} s^x_{j}$. We obtain the final output vector element by accumulating the partial \bfloat{} results as follows:
\begin{equation}
    \label{eq:summation}
    \mathbf{y}_i = \sum_{j=1}^{\ceil{N_c/n}} \mathbf{y}_{ij}^q s^y_{ij}
\end{equation}
ABFP separates scale factors for each row vector $\mathbf{w}_{ij}$ and each column vector $\mathbf{x}_{j}$, and this scheme greatly reduces the quantization effects. By comparison, other schemes include: using only one scale for a submatrix tile \cite{drumond2018training},  one scale for an entire weight tensor \cite{jacob2018quantization, krishnamoorthi2018quantizing}, or separate scales per channel for a given layer (see also \cite{jacob2018quantization, krishnamoorthi2018quantizing}). Typically, in these methods, only one scale is chosen for all input activations to a given layer. In contrast, our method determines the scales of each vector input to the dot product---allowing for a reduced sensitivity to outliers and a smaller quantization error. Employing ABFP does come at the expense of rescaling the dot product inputs, along with additional \bfloat{} storage and multiplications. Importantly, we note that for inference, the weight tensors need only to be converted to and stored in ABFP representation once.

\emph{An aside}: equations~\eqref{eq:quantization} and~\eqref{eq:summation} above clarify why AMS devices suffer more quantization error than digital fixed-point accelerators and why the ABFP representation can remedy that error. In an AMS device, the summation across the different tiles is performed after the final quantization (i.e., after Eq.~\eqref{eq:quantization} and in Eq.~\eqref{eq:summation}). Whereas, the summation across different tiles in a digital fixed-point device can be performed before the final quantization (i.e., in Eq.~\eqref{eq:quantization}).

\subsection{Including Gain in ABFP}
\label{sec:overamp}
Normally, as the tile width ($n$) increases, so do the number of bits required to represent the output of the multiplication (by $\log_2 n$). If the bit precision of the ADC is constant, then more information on the output is lost as fewer lower-significant bits can be captured. However, we discovered that by physically increasing the gain of the analog signal, these lower significant bits can be recovered---at the cost of possibly increased saturation of more significant bits. This effectively increases the precision of the multiplication output and allows DNNs to lose less quality with larger tiles. This is particularly important for DNN acceleration because large tiles allow larger dot products which can be computed within a single clock cycle in an AMS device.

While gain is often used to amplify signals in analog signal acquisition \cite{huang2008front}, it has not to our knowledge been previously employed in the context of BFP numeric representations, adaptive or otherwise.

Within the ABFP numerical format, we can further reduce the quantization error by physically amplifying the analog signal used to compute the analog dot product. Mathematically, it introduces a gain factor $G>1$. This allows for the recovery of less significant bits at the cost of possible saturation of the range of values and a higher energy consumption. Equation~\eqref{eq:quantization} then becomes the following:
\begin{equation}
    \label{eq:quantization_with_gain}
    \mathbf{y}_{ij}^q = Q\left(G \cdot  \mathbf{w}_{ij}^q \cdot \mathbf{x}_{j}^q; n \delta_Y, \tau_Y \right).
\end{equation}
Clamping in the output quantization function becomes necessary because the gain causes some values to overshoot the allowed range. The gain factor is divided out during the accumulation as follows:
\begin{equation}
    \label{eq:summation_with_gain}
    \mathbf{y}_i = \sum_{j=1}^{\ceil{N_c/n}} \mathbf{y}_{ij}^q s^y_{ij}~/~G.
\end{equation}
The intuitive picture can be understood by considering the Eq.~\eqref{eq:quantization_with_gain} above with $G=1$. For the output quantization (typically performed by ADC) not to lose any information, the inequality $n \delta_Y \leq \delta_X \delta_W$ must be satisfied. Phrased differently, the number of bits required to represent the entire output is approximately $b_W + b_X + \log_2 n - 1$. This can easily exceed the bitwidth of today's AMS devices. For example, for $b_W=b_X=8$ and $n=128$ the output is $\approx22$ bits.

\begin{figure} 
\centering
\includegraphics[width=0.48\textwidth]{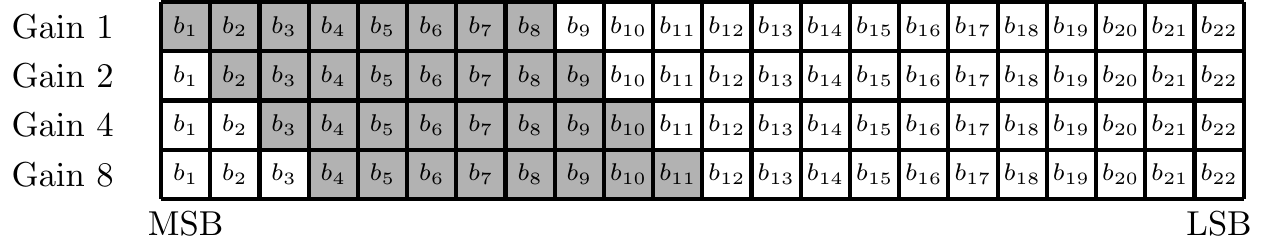}
\caption{Increasing gain allows lower bits to be captured in the output of the ADC while upper bits may saturate. In this example, $b_W=b_X=8$ and $n=128$. The darkened bits are those bit values captured at the output for different gain values.}
\label{fig:gain_as_a_sliding_window}
\vspace{-0.15in}
\end{figure}

Figure~\ref{fig:gain_as_a_sliding_window} illustrates the effect of gain for a fixed $b_W=8$. ABFP mitigates the loss of less-significant bits by increasing the gain of the analog signal, which allows those bits to be recovered. Doubling the analog signal allows one \textit{extra} less-significant bit to be captured at the expense of one \textit{fewer} most-significant bit. It may sound counter-intuitive how trading the higher-order bits for lower-order bits can improve the accuracy of a DNN. However, the distribution of the output of a dot product in a DNN tends to not reach the first few most-significant bits. (See Section \ref{sec:App_error_analysis} of the Appendix for analysis of this aspect of ABFP.)

\subsection{AMS Noise Model in ABFP}
\label{sec:error-model}

The output of a dot product on an AMS device is subject to a certain amount of error in the least significant bits. This error, which is also referred to as ``noise'' in the succeeding sections, is in addition to the existing quantization effects. The origins of this error are stochastic and depend strongly on the physical characteristics of the device. Similarly to Rekhi et al. \cite{rekhi-dally-2019}, we consider it independent of the values of the weights and inputs, and we model this error as a uniformly distributed variable $\mathcal{E}$ with a variance ${\rm Var}(\mathcal{E}) = (n \delta_Y)^2/12$. Therefore, the uniform distribution has a width of one discretization bin~\footnote{It is common to call this discretization bin the least-signficant bit (LSB) as defined in the context of analog signal processing. An LSB is the smallest amount of analog signal that an ADC can discretize, and this corresponds to the width of the output quantization ``bin''. Thus $\pm0.5$ LSB is the maximum quantization error introduced by the ADC.} of the output quantization, and the values that $\mathcal{E}$ takes are between $-n \delta_Y/2$ and $n \delta_Y/2$. The error should be added in Eq.~\eqref{eq:quantization_with_gain} to model the analog dot product as follows:
\begin{equation}
    \label{eq:quantization_with_gain_and_noise}
    \mathbf{y}_{ij}^q = Q\left(G \cdot  \mathbf{w}_{ij}^q \cdot \mathbf{x}_{j}^q + \mathcal{E}; n \delta_Y, \tau_Y  \right).
\end{equation}
The final output vector element is again accumulated as done in Eq.~\eqref{eq:summation_with_gain}. Increasing the gain of the analog signal reduces relative error for outputs that have small magnitudes, albeit at the cost of saturating outputs with larger magnitudes. For the remainder of this paper, ABFP means ABFP with gain and AMS noise.

\section{Finetuning to Recover DNN Quality}
\label{sec:finetuning}

The quantization and noise described in the previous section can affect a model's overall quality. In this section, we discuss two approaches targeted at recovering the original \floatfp{} quality. We discuss an existing approach, namely Quantization-Aware Training or QAT (also called training with simulated quantization \cite{jacob2018quantization}). Next, we present a novel approach to recovering the \floatfp{} quality that we call Differential Noise Finetuning (DNF). Note that we prefer ``finetuning'' over ``training'' to emphasize that DNF starts with a pre-trained DNN and finetunes it.

\begin{figure*}[t]
\centering
\includegraphics[width=0.9\textwidth]{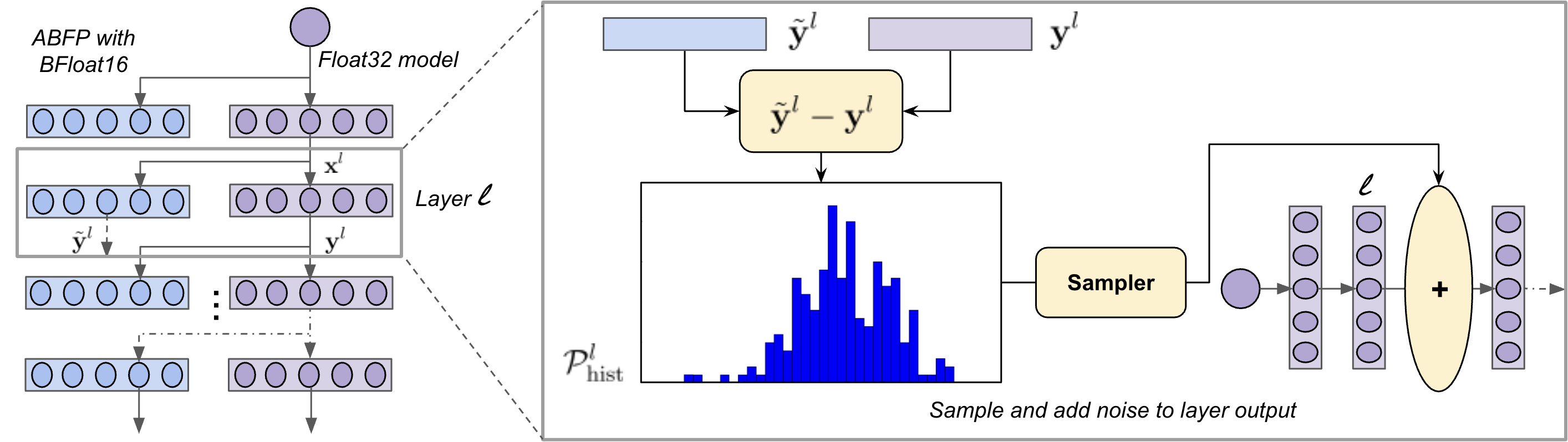}
\caption{Overview of Differential Noise Finetuning (DNF) for a layer $l$. First, the outputs of the \floatfp{} and ABFP DNN layers are computed given the \textit{same} inputs, i.e., output of the previous \floatfp{} layer. Then, the difference between the outputs is converted to a histogram. During finetuning, noise is sampled from the histogram and added to the output of layer $\l$.}
\label{diff_noise_figure}
\end{figure*}

\subsection{Quantization-Aware Training}
QAT simulates quantization and noise effects, i.e., full equation \ref{eq:quantization_with_gain_and_noise}, during the forward pass \cite{jacob2018quantization, krishnamoorthi2018quantizing}. To estimate the gradients of quantization, we use a Straight-Through Estimator (STE). The STE passes the gradient as if the incoming function is an identity function, so its gradient is estimated as $\partial Q(x)/\partial x = 1$.

Let $L$ be the loss function of the DNN computed using ABFP, whose forward pass in a particular layer is defined by Eqs.~\eqref{eq:quantization_with_gain_and_noise} and~\eqref{eq:summation_with_gain}. Gradients during the backward pass are:
\begin{equation}
    \begin{split}
        \frac{\partial L}{\partial \mathbf{x}} =\frac{\partial L}{\partial \mathbf{y}} \cdot \mathbf{W} \; \text{and} \; 
        \frac{\partial L}{\partial \mathbf{W}} = \mathbf{x} \cdot \frac{\partial L}{\partial \mathbf{y}}
    \end{split}    
\end{equation}
The gradients during the backward pass are accumulated in \floatfp{}. Hence QAT is a case of mixed-precision training. We start performing QAT from a pre-trained DNN. Once training is complete, we run inference with ABFP.

\subsection{Differential Noise Finetuning}
\label{diff_noise}
DNF enables DNNs to recover from quality lost due to noise and quantization. Unlike QAT, DNF retains \floatfp{}
precision in the forward pass (without tiling or quantization which can be slow in standard digital hardware, e.g., CPU or GPU), and adds noise to the outputs of the layers of the \floatfp{} model during finetuning. The added noise enables the DNN to effectively adapt to quantization error and noise. To achieve this, the added noise is sampled from a histogram of the differences between the ABFP and \floatfp{} model layer outputs. We refer to these differences as ``differential noise''. By sampling additive differential noise this way, DNF emulates the aggregate effects of quantization error and noise.

Figure \ref{diff_noise_figure} describes the key steps in DNF. We begin by computing the output differences between the \floatfp{} and the ABFP layers. The input $\mathbf{x}^l$ to the \floatfp{} and ABFP layers $l$ is the \textit{same}: the output of the previous \floatfp{} layer. Further, we denote $f^l(\cdot)$ as a layer operation in the \floatfp{} DNN, and $\tilde{f}^l(\cdot)$ as the same layer operation in the ABFP DNN. Similarly, we denote the \floatfp{} and the ABFP layer outputs as $\mathbf{y}^l$ and $\tilde{\mathbf{y}}^l$, respectively: $\mathbf{y}^l = f^l(\mathbf{x}^l)$ and  $\tilde{\mathbf{y}}^l = \tilde{f}^l(\mathbf{x}^l)$. Given the same inputs, we then compute the corresponding output differences as $ \Delta{\mathbf{y}}^l = \tilde{\mathbf{y}}^l - \mathbf{y}^l$. Since the output differences $\Delta \mathbf{y}^l$ are computed for every layer, DNF captures the quantization and noise effects at the layer-level. Once computed, $\Delta \mathbf{y}^l$ is used to instantiate per-layer differential noise distributions. In particular, the values in $\Delta \mathbf{y}^l$ are converted into a histogram. The computed histograms are smoothed~\footnote{We add 0.5 to each histogram bin to avoid zero probabilities.} and then used to approximate a probability distribution for each layer $l$, denoted by $\mathcal{P}^l_{\text{hist}}$. 

During finetuning, a differential noise tensor $\boldsymbol{\xi}^l$, sampled from the constructed histogram, is added in the forward pass:
\begin{equation}
\label{eq:differential_noise_sampling}
    \mathbf{y}^l = f^l(\mathbf{x}^l) + \boldsymbol{\xi}^l,
    \text{where } \boldsymbol{\xi}^l \sim \mathcal{P}^l_{\text{hist}}.
\end{equation}
Note that the gradients during the backward pass are accumulated in \floatfp{}. Computationally, the amount of overhead of sampling and adding depends on two key parameters: 
\begin{inparaenum}[(1)]
\item the dimension of the noise tensor, and
\item the number of bins used within the histogram representation.
\end{inparaenum}
The histogram for each layer only needs to be computed \textit{once} before finetuning begins. Hence, the key overhead during finetuning is the time taken to sample from a histogram, which is proportional to the number of bins and noise size.

In principle, $\Delta \mathbf{y}^l$ can be statistically analyzed to estimate differential noise present at each layer, e.g., we can compute the mean and standard deviations of $\Delta \mathbf{y}^l$. A non-zero mean shows a bias introduced by the ABFP number representaton, and a larger standard deviation indicates a stronger susceptibility to quantization error and noise.

\begin{figure*}[t]
\centering
\includegraphics[width=1.0\textwidth]{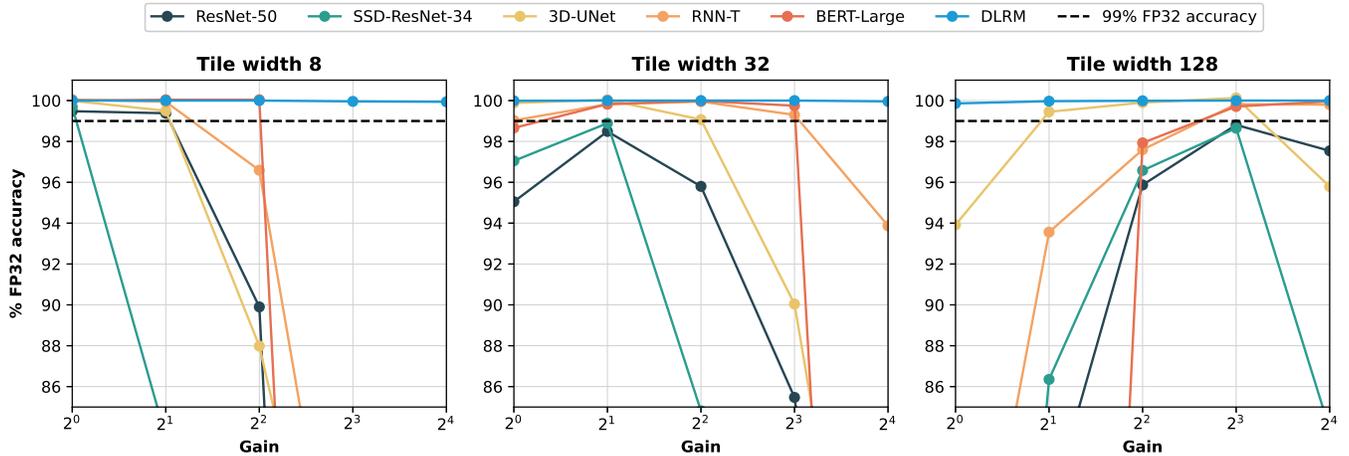}
\vspace{-0.4in}
\caption{Model metrics as a percent of \floatfp{} accuracy relative to gain at $b_W/b_X/b_Y = 8/8/8$. The black dashed-line is the 99\% threshold. All models drop less than 1\% \floatfp{} in accuracy at certain tile width and gain combinations.}
\label{fig:score_plots}
\end{figure*}

\begin{table*}[t]
\centering
\caption{\label{table:mlperf} Model metrics for different tile widths, gains and bitwidths. Results in bold are above 99\% of the \floatfp{} result. The metrics for the models are top-1 accuracy (ResNet50), mean average precision or mAP score (SSD-ResNet34), mean accuracy (3D U-Net), accuracy ($1 - \text{WER}$) (RNN-T), F1 score (BERT-Large), and ROC AUC (DLRM).}
\vspace{0.1in}
\begin{tabular}{r c|c c c c c |c c c c c}

 \multicolumn{2}{c}{} & \multicolumn{5}{c}{$b_W/b_X/b_Y = 6/6/8$} & \multicolumn{5}{c}{$b_W/b_X/b_Y = 8/8/8$} \\ \hline
 \multicolumn{2}{c|}{\bf ResNet50} &  \multicolumn{5}{c|}{\bf Gain} & \multicolumn{5}{c}{\bf Gain} \\
 \multicolumn{2}{c|}{\floatfp{}: 76.13} & 1 & 2 & 4 & 8 & 16 & 1 & 2 & 4 & 8 & 16 \\ \hline
 \multirow{3}{*}{\rotatebox[origin=l]{90}{\parbox{12mm}{\centering \bf Tile \\ width}}} &
    8 & \textbf{75.59} & \textbf{75.58} & 68.69 & 0.65 & 0.10 & \textbf{75.74} & \textbf{75.65} & 68.44 & 0.65& 0.10 \\
 & 32 & 71.98 & 74.60 & 72.55 & 64.63 & 40.47 & 72.36 & 74.98 & 72.93 & 65.07 & 39.76 \\
 & 128 & 0.57 & 59.71 & 72.42 & 74.69 & 73.24 & 0.66 & 60.80 & 72.98 & 75.23 & 74.26 \\ \hline \hline

 \multicolumn{2}{c|}{\bf SSD-ResNet34} &  \multicolumn{5}{c|}{\bf Gain} & \multicolumn{5}{c}{\bf Gain} \\
 \multicolumn{2}{c|}{\floatfp{}: 19.59} & 1 & 2 & 4 & 8 & 16 & 1 & 2 & 4 & 8 & 16 \\ \hline
 \multirow{3}{*}{\rotatebox[origin=l]{90}{\parbox{12mm}{\centering \bf Tile \\ width}}} &
    8 & \textbf{19.47} &  16.28 & 6.17 & 0.00 & 0.00 & \textbf{19.53} & 16.37 & 6.16 & 0.00 & 0.00 \\
 & 32 & 18.91 & 19.26 & 16.64 & 10.97 & 0.46 & 19.01 & 19.37 & 16.61 & 10.95 & 0.47 \\
 & 128 & 7.00 & 16.77 & 18.78 & 19.16 & 16.36 & 7.08 & 16.92 & 18.92 & 19.32 & 16.44 \\ \hline \hline
 
 \multicolumn{2}{c|}{\bf 3D U-Net} &  \multicolumn{5}{c|}{\bf Gain} & \multicolumn{5}{c}{\bf Gain} \\
 \multicolumn{2}{c|}{\floatfp{}: 85.30} & 1 & 2 & 4 & 8 & 16 & 1 & 2 & 4 & 8 & 16 \\ \hline
 \multirow{3}{*}{\rotatebox[origin=l]{90}{\parbox{12mm}{\centering \bf Tile \\ width}}} &
    8 & \textbf{85.31} &  \textbf{84.89} & 75.01 & 58.04 & 50.38 & \textbf{85.29} & \textbf{84.88} & 75.05 & 58.19 & 50.43 \\
 & 32 & \textbf{85.29} & \textbf{85.40} & \textbf{84.62} & 77.08 & 54.24 & \textbf{85.24} & \textbf{85.33} & \textbf{84.52} & 76.80 & 53.74 \\
 & 128 & 80.03 & \textbf{84.87} & \textbf{85.29} & \textbf{85.40} & 81.67 & 80.11 & \textbf{84.65} & \textbf{85.22} & \textbf{85.41} & 81.72 \\ \hline \hline

 \multicolumn{2}{c|}{\bf RNN-T} &  \multicolumn{5}{c|}{\bf Gain} & \multicolumn{5}{c}{\bf Gain} \\
 \multicolumn{2}{c|}{\floatfp{}: 92.55} & 1 & 2 & 4 & 8 & 16 & 1 & 2 & 4 & 8 & 16 \\ \hline
 \multirow{3}{*}{\rotatebox[origin=l]{90}{\parbox{12mm}{\centering \bf Tile \\ width}}} &
    8 & \textbf{92.49} &  \textbf{92.47} & 89.29 & 64.69 & 0.00 & \textbf{92.60} & \textbf{92.50} & 89.40 & 64.80 & 0.00 \\
 & 32 & \textbf{91.94} & \textbf{92.39} & \textbf{92.55} & \textbf{91.96} & 86.83 & \textbf{91.65} & \textbf{92.39} & \textbf{92.51} & \textbf{91.90} & 86.88 \\
 & 128 & 65.81 & 89.93 & 91.44 & \textbf{91.81} & \textbf{91.80} & 63.97 & 86.59 & 90.33 & \textbf{92.39} & \textbf{92.37} \\ \hline \hline
 
 \multicolumn{2}{c|}{\bf BERT-Large} &  \multicolumn{5}{c|}{\bf Gain} & \multicolumn{5}{c}{\bf Gain} \\
 \multicolumn{2}{c|}{\floatfp{}: 93.15} & 1 & 2 & 4 & 8 & 16 & 1 & 2 & 4 & 8 & 16 \\ \hline
 \multirow{3}{*}{\rotatebox[origin=l]{90}{\parbox{12mm}{\centering \bf Tile \\ width}}} &
    8 & \textbf{93.17} &  \textbf{93.23} & \textbf{93.18} & 9.17 & 6.94  & \textbf{93.15} & \textbf{93.19} & \textbf{93.19} & 9.14 &  6.99 \\
 & 32 & 91.74 & \textbf{92.92} & \textbf{93.16} & \textbf{92.91} & 17.38 & 91.90 & \textbf{93.00} & \textbf{93.14} & \textbf{92.92} & 17.23 \\
 & 128 & 5.10 & 4.34 & 90.82 & \textbf{92.70} & \textbf{92.89} & 5.37 & 4.59 & 91.23 & \textbf{92.88} & \textbf{93.12} \\ \hline \hline
 
 \multicolumn{2}{c|}{\bf DLRM} &  \multicolumn{5}{c|}{\bf Gain} & \multicolumn{5}{c}{\bf Gain} \\
 \multicolumn{2}{c|}{\floatfp{}: 80.35} & 1 & 2 & 4 & 8 & 16 & 1 & 2 & 4 & 8 & 16 \\ \hline
 \multirow{3}{*}{\rotatebox[origin=l]{90}{\parbox{12mm}{\centering \bf Tile \\ width}}} &
    8 & \textbf{80.35} &  \textbf{80.35} & \textbf{80.35} & \textbf{80.32} & \textbf{80.30} & \textbf{80.35} & \textbf{80.35} & \textbf{80.35} & \textbf{80.32} & \textbf{80.30} \\
 & 32 & \textbf{80.34} & \textbf{80.35} & \textbf{80.35} & \textbf{80.35} & \textbf{80.32} & \textbf{80.34} & \textbf{80.35} & \textbf{80.35} & \textbf{80.35} & \textbf{80.32} \\
 & 128 & \textbf{80.24} & \textbf{80.32} & \textbf{80.34} & \textbf{80.35} & \textbf{80.35} & \textbf{80.24} & \textbf{80.32} & \textbf{80.34} & \textbf{80.35} & \textbf{80.35} \\ \hline
 
\end{tabular}
\end{table*}

\section{Results}
\label{results}

\begin{figure*}[t]
\includegraphics[width=\textwidth]{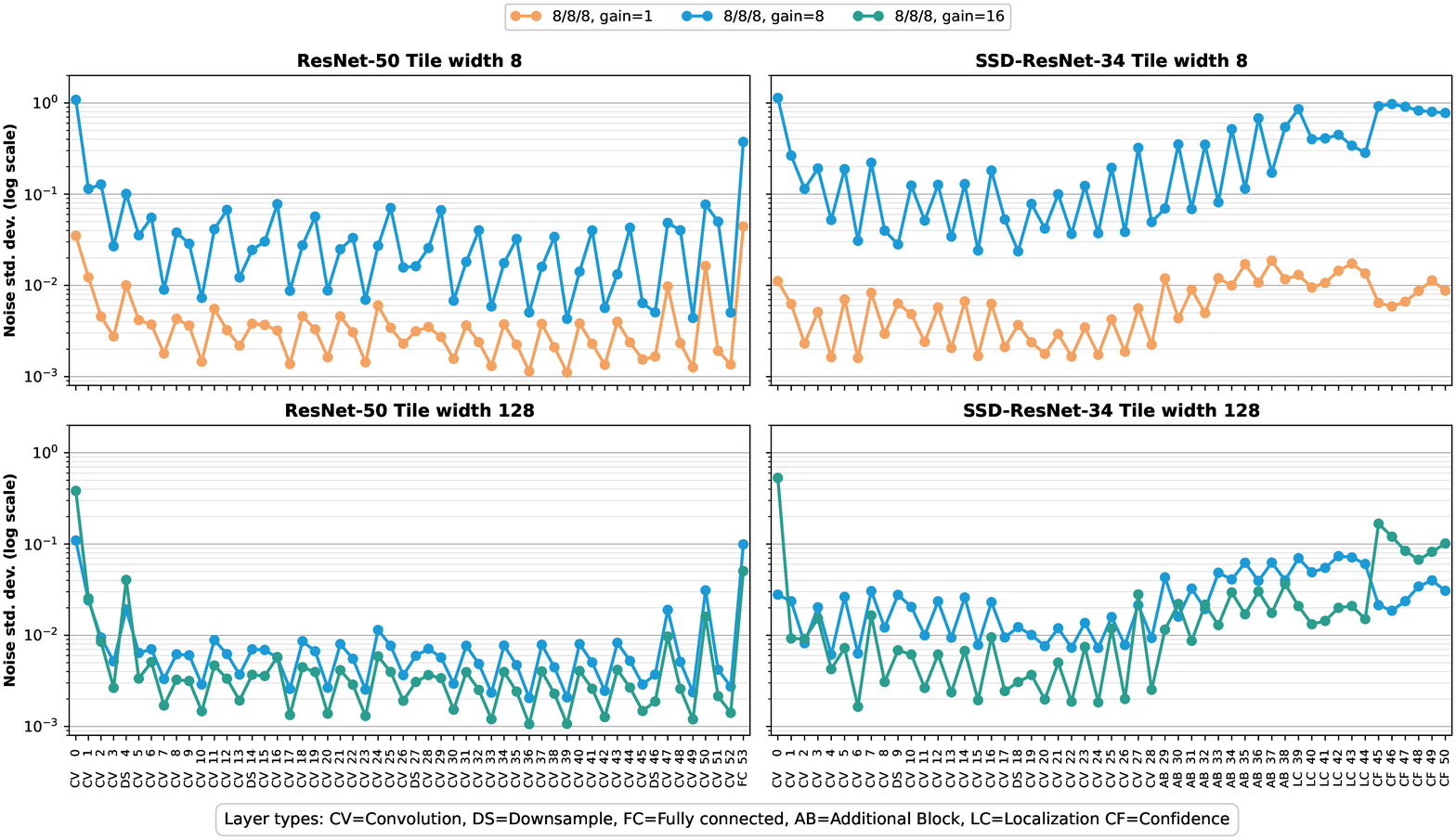}
\vspace{-0.2in}
\caption{Differential noise standard deviations for a subset of layers in ResNet50 and SSD-ResNet34, at tile widths 128 (top) and 8 (bottom) and $b_W/b_X/b_Y = 8/8/8$. At tile width 8, lower gain corresponds to a lower noise standard deviation and a higher model quality. At tile width 128, all layers do not exhibit the same sensitivity to increased gain. The first layer of ResNet50 and the first and last few layers of SSD-ResNet34 exhibit much higher noise standard deviations at gain of 16 than at gain of 8, resulting in a drop in accuracy.}
\label{fig:noise_std_resnet_ssd_1}
\end{figure*}

In this section, we present results of running the MLPerf\texttrademark{} datacenter inference benchmark DNNs using ABFP with ADC noise. We simulated ABFP with custom Pytorch layers that take \bfloat{} input tensors and output \bfloat{} tensors. During inference and when finetuning with QAT, the convolutions of ResNet50, SSD-ResNet34, and 3D-UNet are converted to tiled matrix-multiplications using the im2col algorithm \cite{chellapilla2006im2col}. Other layers, such as batch-norm, layer-norm, pooling, and non-linearities (e.g., tanh, softmax, and GELU), read \bfloat{} inputs from memory and then perform arithmetic in \floatfp{} before converting back to \bfloat{} as done by Micikevicius et al.~\cite{micikevicius2017mixed}. For ResNet50, the reported results use batch-norm folding although there is no significant difference with or without using batch-norm folding during inference.

\subsection{Inference before Finetuning}
\label{sec:inference_before_finetuning}

We evaluate ABFP on all six MLPerf\texttrademark{} datacenter inference benchmark DNNs \cite{reddi2020mlperf} using pre-trained PyTorch checkpoints (refer Table~\ref{tab:ml_perf_links} in the appendix for the checkpoints). For each DNN, we run inference over tile widths $\{8, 32, 128\}$, gains $\{1, 2, 4, 8, 16\}$, and bitwidths $b_W/b_X/b_Y$ $\{6/6/8, 8/8/8\}$. The gain increases by powers of 2 because each doubling of gain captures an additional less-significant bit (see Figure~\ref{fig:gain_as_a_sliding_window}). Table~\ref{tab:ml_commons_tasks} lists the datasets on which each DNN was evaluated. The average of all the runs is reported in Table~\ref{table:mlperf} (cf. Table~\ref{table:mlperf_std} in the appendix for the standard deviations of the metrics in Table~\ref{table:mlperf}). With the exception of 3D U-Net, we evaluate each DNN 10 times. We evaluate 3D U-Net three times because of computational constraints; the average across its three runs is reported.

Table~\ref{table:mlperf} shows that the quality of each DNN varies depending on tile width and gain. At tile width 8, all DNNs perform with little loss of \floatfp{} accuracy at gain of 1, and the loss increases with increasing gain. The opposite is the case at tile width 128, where quality generally improves with increasing gain (cf. section \ref{sec:App_error_analysis} in appendix). For a given tile width, gain controls the tradeoff between saturation and precision that is important for DNN quality. Figure~\ref{fig:score_plots} highlights that all DNNs lose less than 1\% of \floatfp{} quality at certain tile width and gain combinations. 

Changing the input and weight bitwidths from 8 to 6 bits has little effect on DNN quality, especially compared to the effects of changing tile width. SSD-ResNet34 and ResNet50 have a larger decrease in quality as compared to other DNNs, perhaps due to having a regression component, or a high number of output classes.

Predictions about DNN quality cannot be made on tile width or gain alone. Hence, a layer-wise analysis of the differential noise (see Eq.~\eqref{eq:differential_noise_sampling}) can be useful. Figure \ref{fig:noise_std_resnet_ssd_1} shows noise standard deviations for ResNet50 and SSD-ResNet34. At tile width 8, a lower gain corresponds to lower noise standard deviations for \textit{all} the layers and to higher quality for both DNNs, as shown in Table \ref{table:mlperf}. However, at tile width 128 and gain of 16, all layers do not exhibit the same noise response. For ResNet50, the first layer has a much larger noise standard deviation at gain 16 than at gain 8 (standard deviations for the other layers are quite close at both gains), leading to a minor drop of 1\% accuracy at gain 16. For SSD-ResNet34, the first layer \textit{and} the last confidence layers have much larger noise standard deviations at gain 16 than at gain 8, leading to a larger drop of about 3\% in mAP. Performance at the 6/6/8 bitwidths is shown in Appendix Figure \ref{fig:noise_std_resnet_ssd_app} (note that noise standard deviations and corresponding DNN qualities between 6/6/8 and 8/8/8 are very close). A layer-wise analysis of the DNNs provides insights into layers that are more susceptible to quantization error and noise. We leverage the insights in the next section to tailor the finetuning of ResNet50 and SSD-ResNet34.

\subsection{Inference after Finetuning}
\label{sec:inf_after_finetuning}
We demonstrate the effectiveness of QAT and DNF at tile width 128 and gain of 8 on ResNet50 and SSD-ResNet34: the only two DNNs that fall just below 99\% of their respective \floatfp{} quality at this setting. We choose to investigate retraining at tile width 128, since this tile width enables an AMS device to run the fastest, in comparison to the rest of the studied tile widths\footnote{An AMS device with a tile width $n$ is capable of performing a dot product of vectors of length $n$, with a single clock cycle. Typically, AMS devices are architected to perform matrix-vector multiplications. An AMS device with a matrix tile dimension of $n \times n$ is able to perform a multiplication between an $n \times n$ matrix and an $n$-long vector in a \textit{single} clock cycle.}.

We finetune ResNet50 with QAT and DNF separately using the AdamW optimizer~\cite{DBLP:journals/corr/abs-1711-05101} with a learning rate of \num{1e-6}. This learning rate decreases multiplicatively by a factor 0.3 per epoch. We run QAT for 2 epochs with a batch size of 100 and DNF for 5 epochs with a batch size of 128. We finetune SSD-ResNet34 with QAT and DNF using the SGD optimizer with learning rates of \num{1e-6} and \num{2.169e-5} respectively. We use a cosine annealing, one-cycle learning rate scheduler. The SGD optimizer uses weight-decay of \num{5e-4} and momentum of 0.728. We run both QAT and DNF for 8 epochs, with batch sizes of 4 and 24 respectively. 

During QAT with ResNet50, the batch-norm layers are folded. Since there is no significant difference in quality with or without batch-norm folding, we report results with folding. When finetuning SSD-ResNet34, batch-norm layers are not folded. The backward pass is performed fully in \floatfp{}.

When finetuning SSD-ResNet34, DNF is slower than conventional training in \floatfp{} due to the time cost of sampling from a histogram (see Eq.~\eqref{eq:differential_noise_sampling}). To minimize this cost, we add differential noise only to those layers with the highest noise standard deviations, as higher variance suggests greater susceptibility to quantization error and noise. Figure \ref{fig:noise_std_resnet_ssd_1}~shows that the deeper layers of SSD-ResNet34 (including localization and confidence) have higher noise standard deviations. Hence, when finetuning we add differential noise only to these layers. In the case of ResNet50, we add differential noise to all the convolutional and linear layers, since it does not incur a large sampling overhead.
This is in contrast to previous approaches in adapting DNNs to AMS hardware, which do not add noise to the first and last layers of ResNet50~\cite{rekhi-dally-2019}. Our DNF noise histograms are normalized $100$-bin histograms created from sampling \textit{one} batch of data with sizes 128 and 24 for ResNet50 and SSD-ResNet34, respectively.

\begin{table}[h]
\vspace{-0.15in}
\centering
\caption{\label{table:retraining} Comparing QAT and DNF on ResNet50 and SSD-ResNet34. Bold values highlight methods that improved the DNN to above 99\% of the original \floatfp{} metric. The metrics are top-1 accuracy (ResNet50) and mAP (SSD-ResNet34).}
\vspace{0.1in}
\begin{tabular}{cccc}
 & & \multicolumn{2}{c}{$b_W/b_X/b_Y$} \\ \cline{3-4}
DNN & Finetuning & $6/6/8$ & $8/8/8$ \\ \hline
\multirow{3}{*}{\bf ResNet50} & \multicolumn{1}{l}{No finetuning} & 74.69 & 75.23 \\
 & \multicolumn{1}{l}{QAT} & \bf 76.04 & \bf 76.14 \\
 & \multicolumn{1}{l}{DNF} &  75.15  & \bf 75.97  \\ \hline
\multirow{3}{*}{\bf SSD-ResNet34} & \multicolumn{1}{l}{No finetuning} & 19.16 & 19.32 \\
 & \multicolumn{1}{l}{QAT} & 19.23 & 19.34 \\
 & \multicolumn{1}{l}{DNF} &  \bf 19.55  & \bf 19.64  \\ \hline
\end{tabular}
\end{table}

Table \ref{table:retraining} shows the quality metrics of the DNNs in ABFP after finetuning them with QAT or DNF. With QAT, ResNet50 \floatfp{} accuracy is recovered within 2 epochs; DNF improves the DNN accuracy towards 99\% of the \floatfp{} accuracy at bitwidths $8/8/8$ within 5 epochs. DNF fully recovers the \floatfp{} mAP of SSD-ResNet34 within 8 epochs, while QAT fails to achieve 99\% of the \floatfp{} mAP in both bitwidth configurations. 

QAT was about $4 \times$ slower than DNF with ResNet50 and SSD-ResNet34 using an NVIDIA A-100 GPU. DNF is much faster than QAT with ABFP, because QAT has to perform the full tiling, scaling, and error-sampling operations associated with simulating ABFP in a digital hardware. On the other hand, DNF only has to sample from the differential noise histogram. Also, analyzing the layer-wise noise statistics allows DNF to be tailored to specific DNNs---unlike vanilla QAT---thus further improving computational speed.

\section{Discussion}
\label{discussion}

The results in the previous section show that there are optimal choices of gain and tile width for different DNNs. It can be advantageous for an AMS hardware to be able to change gain or tile-width. Changing the tile-width can be difficult as the size of the analog matrix multiplication unit may already be fixed in the hardware. However, changing the gain may be more easily done by either amplifying or attenuating the analog signals that are used for the multiplication.

As a numerical representation, ABFP is highly suitable for AMS hardware. Generally the activation tensors can have a wide range of values which can lead to significant loss of information if the range is too wide. Some DNN operations, e.g., the batch-norm or layer-norm operation, are sensitive to both the small and large values of the tensor. Thus, in AMS hardware, such operations should be performed in digital floating-point representation. Dot products, however, do not have this issue, because the elements with the largest magnitudes dominate the summation, while elements with small magnitudes have little impact on the final result. As such, dot-products are well suited for AMS hardware, whose physics constraints require the input and the output data to have a similar range.

One may think that the power savings afforded by the AMS hardware is diminished by the repeated conversions between \bfloat{} and the ABFP, before and after the matrix multiplications. However, the overhead of the conversions is amortized over the reduction within the matrix multiplication. For example, in a (naive) matrix-matrix multiplication between two $N \times N$ matrices with $2N^3$ multiply-and-accumulate operations, only a total of $2N^2/n$ \bfloat{} to ABFP conversions are necessary (where $n$ is the tile width). Furthermore, for the weight matrix of a DNN, the conversion only needs to happen once before any inference occurs. The computational cost of the scales of the ABFP can also be further reduced by restricting the scales to be exponents only, without any mantissa---albeit with possible loss of some numerical precision.

Our results on the MLPerf\texttrademark{} datacenter inference DNNs directly address the accuracy-energy tradeoff curve of Rekhi et~al.~\cite{rekhi-dally-2019}. According to their findings at tile width of 8, using 12.5 ADC bits achieved accuracy loss consistent with zero for ResNet50. Our method, using tile width of 128, a gain of 8, and the same noise model and accuracy constraint, needs only 8 output bits (see second row of Table~\ref{table:retraining}). The energy savings from reducing the ADC bits is $2^{12.5-8} \approx 23\times$, while the energy increase with a gain of 8 is a factor of 8$\times$, so overall our method reduces energy by a factor of $\approx2.8$ compared with the optimal design configuration in Rekhi et~al.~\cite{rekhi-dally-2019} for ResNet50. We also stress that an AMS device with a tile width of 128 executes $16\times$ more multiply-accumulate operations per clock cycle than that with a tile width of 8. For future work, we believe it would be helpful to have an exact, holistic model for the energy usage, inference runtime, and accuracy implications of the AMS design space. For example, the above analysis considers only the ADC's energy contribution. The energy vs. accuracy analysis that is done by Rekhi et al.~\cite{rekhi-dally-2019} could be further extended to incorporate our ABFP representation. 

With respect to accuracy, recent work has shown that pruning or quantizing an image-classification DNN can cause it to change predictions on pruning-identifiable exemplars (PIE). A PIE is a mislabeled or otherwise a very difficult-to-classify DNN input \cite{hooker2021compressed}. Indeed, we saw DNNs with few output classes (e.g., DLRM with 2 classes, 3D-UNet with 2 classes, RNN-T with 29 classes) tend to retain high quality under more ABFP configurations than DNNs with many output classes (e.g., ResNet50 with 1000 classes). This suggests that output dimensionality and decision-boundary complexity influence a DNN's robustness to AMS quantization error and noise.

Thus far, we have used the noise $\mathcal{E}$ to model the aggregate noise (e.g., thermal noise, shot noise, multiplier nonlinearity, and ADC quantization noise) of the AMS device. However, the exact models of the noise may depend on the architecture of the analog hardware, and they may also depend on the inputs/weights. We plan to investigate the effect of ABFP and our finetuning approaches against the specific hardware architectures. Our finetuning approaches---QAT and DNF---are generic enough to accommodate different magnitudes and models of the noise.

Lastly, we can incorporate techniques from quantization for limited-precision digital hardware~\cite{gholami2021quant_survey}. For example, we can use measured percentiles for scaling the ABFP numbers~\cite{mckinstry2019percentile,haowu2020quantization} (instead of max values) or the DNF can use per-channel noise/error distribution~\cite{welling_2019_ICCV_channelwiseQ,jacob2018quantization}. However, these are beyond the scope of our current work, and will be explored in the future. We will also investigate the error caused by quantization, tiling, gain and stochastic noise to construct more compact noise distributions and improve DNN quality.

\section{Conclusion}
\label{conclusion}
Innovative AMS devices offer faster inference with a lower-power footprint by comparison with traditional digital electronics. However, the lower precision typically used on these devices and the presence of stochastic error require the application of specialized methods for numerical representation to achieve results at the level of \floatfp{} precision. Here we introduced such a method---ABFP with gain---which allows for lower quantization effects when the tile size of the device is made large in order to speed up the matrix-matrix multiplication. We also showed that in the cases where some DNNs do not achieve the required quality without finetuning, a fast new training method called Differential noise finetuning (DNF) restores the DNN quality. Thus, we achieve less than 1\% loss in model quality for the six MLPerf\texttrademark{} datacenter inference benchmark DNNs.

\bibliography{nnls}

\begin{thebibliography}{10}
\providecommand{\url}[1]{#1}
\csname url@samestyle\endcsname
\providecommand{\newblock}{\relax}
\providecommand{\bibinfo}[2]{#2}
\providecommand{\BIBentrySTDinterwordspacing}{\spaceskip=0pt\relax}
\providecommand{\BIBentryALTinterwordstretchfactor}{4}
\providecommand{\BIBentryALTinterwordspacing}{\spaceskip=\fontdimen2\font plus
\BIBentryALTinterwordstretchfactor\fontdimen3\font minus
  \fontdimen4\font\relax}
\providecommand{\BIBforeignlanguage}[2]{{%
\expandafter\ifx\csname l@#1\endcsname\relax
\typeout{** WARNING: IEEEtran.bst: No hyphenation pattern has been}%
\typeout{** loaded for the language `#1'. Using the pattern for}%
\typeout{** the default language instead.}%
\else
\language=\csname l@#1\endcsname
\fi
#2}}
\providecommand{\BIBdecl}{\relax}
\BIBdecl

\bibitem{patterson2021carbon}
D.~Patterson, J.~Gonzalez, Q.~Le, C.~Liang, L.-M. Munguia, D.~Rothchild, D.~So,
  M.~Texier, and J.~Dean, ``Carbon emissions and large neural network
  training,'' \emph{arXiv preprint arXiv:2104.10350}, 2021.

\bibitem{dlinference_fb}
\BIBentryALTinterwordspacing
J.~Park, M.~Naumov, P.~Basu, S.~Deng, A.~Kalaiah, D.~S. Khudia, J.~Law,
  P.~Malani, A.~Malevich, N.~Satish, J.~M. Pino, M.~Schatz, A.~Sidorov,
  V.~Sivakumar, A.~Tulloch, X.~Wang, Y.~Wu, H.~Yuen, U.~Diril, D.~Dzhulgakov,
  K.~M. Hazelwood, B.~Jia, Y.~Jia, L.~Qiao, V.~Rao, N.~Rotem, S.~Yoo, and
  M.~Smelyanskiy, ``Deep learning inference in facebook data centers:
  Characterization, performance optimizations and hardware implications,''
  \emph{CoRR}, vol. abs/1811.09886, 2018. [Online]. Available:
  \url{http://arxiv.org/abs/1811.09886}
\BIBentrySTDinterwordspacing

\bibitem{dally2015high}
W.~Dally, ``High-performance hardware for machine learning,'' \emph{NIPS
  Tutorial}, vol.~2, 2015.

\bibitem{seok2019cases}
M.~Seok, M.~Yang, Z.~Jiang, A.~A. Lazar, and J.-S. Seo, ``Cases for analog
  mixed signal computing integrated circuits for deep neural networks,'' in
  \emph{2019 International Symposium on VLSI Design, Automation and Test
  (VLSI-DAT)}.\hskip 1em plus 0.5em minus 0.4em\relax IEEE, 2019, pp. 1--2.

\bibitem{deng2020model}
L.~Deng, G.~Li, S.~Han, L.~Shi, and Y.~Xie, ``Model compression and hardware
  acceleration for neural networks: A comprehensive survey,'' \emph{Proceedings
  of the IEEE}, vol. 108, no.~4, pp. 485--532, 2020.

\bibitem{rekhi-dally-2019}
\BIBentryALTinterwordspacing
A.~S. Rekhi, B.~Zimmer, N.~Nedovic, N.~Liu, R.~Venkatesan, M.~Wang,
  B.~Khailany, W.~J. Dally, and C.~T. Gray, ``Analog/mixed-signal hardware
  error modeling for deep learning inference,'' in \emph{Proceedings of the
  56th Annual Design Automation Conference 2019}, ser. DAC '19.\hskip 1em plus
  0.5em minus 0.4em\relax New York, NY, USA: Association for Computing
  Machinery, 2019. [Online]. Available:
  \url{https://doi.org/10.1145/3316781.3317770}
\BIBentrySTDinterwordspacing

\bibitem{ghodrati2020}
\BIBentryALTinterwordspacing
S.~Ghodrati, H.~Sharma, S.~Kinzer, A.~Yazdanbakhsh, J.~Park, N.~S. Kim,
  D.~Burger, and H.~Esmaeilzadeh, ``Mixed-signal charge-domain acceleration of
  deep neural networks through interleaved bit-partitioned arithmetic,'' in
  \emph{Proceedings of the ACM International Conference on Parallel
  Architectures and Compilation Techniques}, ser. PACT '20.\hskip 1em plus
  0.5em minus 0.4em\relax New York, NY, USA: Association for Computing
  Machinery, 2020, p. 399–411. [Online]. Available:
  \url{https://doi.org/10.1145/3410463.3414634}
\BIBentrySTDinterwordspacing

\bibitem{reddi2020mlperf}
V.~J. Reddi, C.~Cheng, D.~Kanter, P.~Mattson, G.~Schmuelling, C.-J. Wu,
  B.~Anderson, M.~Breughe, M.~Charlebois, W.~Chou \emph{et~al.}, ``Mlperf
  inference benchmark,'' in \emph{2020 ACM/IEEE 47th Annual International
  Symposium on Computer Architecture (ISCA)}.\hskip 1em plus 0.5em minus
  0.4em\relax IEEE, 2020, pp. 446--459.

\bibitem{visionmlperf2021}
V.~J. Reddi, C.~Cheng, D.~Kanter, P.~Mattson, G.~Schmuelling, and C.-J. Wu,
  ``The vision behind mlperf: Understanding ai inference performance,''
  \emph{IEEE Micro}, vol.~41, no.~3, pp. 10--18, 2021.

\bibitem{bucilua2006model}
C.~Bucilua, R.~Caruana, and A.~Niculescu-Mizil, ``Model compression,'' in
  \emph{Proceedings of the 12th ACM SIGKDD international conference on
  Knowledge discovery and data mining}, 2006, pp. 535--541.

\bibitem{hinton2015distilling}
G.~Hinton, O.~Vinyals, and J.~Dean, ``Distilling the knowledge in a neural
  network,'' \emph{arXiv preprint arXiv:1503.02531}, 2015.

\bibitem{lecun1990optimal}
Y.~LeCun, J.~S. Denker, and S.~A. Solla, ``Optimal brain damage,'' in
  \emph{Advances in neural information processing systems}, 1990, pp. 598--605.

\bibitem{blalock2020state}
D.~Blalock, J.~J.~G. Ortiz, J.~Frankle, and J.~Guttag, ``What is the state of
  neural network pruning?'' \emph{arXiv preprint arXiv:2003.03033}, 2020.

\bibitem{liberty2013simple}
E.~Liberty, ``Simple and deterministic matrix sketching,'' in \emph{Proceedings
  of the 19th ACM SIGKDD international conference on Knowledge discovery and
  data mining}, 2013, pp. 581--588.

\bibitem{dasgupta2010sparse}
A.~Dasgupta, R.~Kumar, and T.~Sarlós, ``A sparse johnson: Lindenstrauss
  transform,'' in \emph{Proceedings of the forty-second ACM symposium on Theory
  of computing}, 2010, pp. 341--350.

\bibitem{abadi2016tensorflow}
M.~Abadi, A.~Agarwal, P.~Barham, E.~Brevdo, Z.~Chen, C.~Citro, G.~S. Corrado,
  A.~Davis, J.~Dean, M.~Devin \emph{et~al.}, ``Tensorflow: Large-scale machine
  learning on heterogeneous distributed systems,'' \emph{arXiv preprint
  arXiv:1603.04467}, 2016.

\bibitem{khudia2018open}
\BIBentryALTinterwordspacing
D.~S. Khudia, P.~Basu, and S.~Deng, ``Open-sourcing fbgemm for state-of-the-art
  server-side inference,'' 2018. [Online]. Available:
  \url{https://engineering.fb.com/2018/11/07/ml-applications/fbgemm/}
\BIBentrySTDinterwordspacing

\bibitem{nvidia2017v100}
T.~NVIDIA, ``V100 gpu architecture,'' \emph{The world's most advanced data
  center GPU. Version WP-08608-001\_v1}, vol.~1, 2017.

\bibitem{jouppi2017datacenter}
N.~P. Jouppi, C.~Young, N.~Patil, D.~Patterson, G.~Agrawal, R.~Bajwa, S.~Bates,
  S.~Bhatia, N.~Boden, A.~Borchers \emph{et~al.}, ``In-datacenter performance
  analysis of a tensor processing unit,'' in \emph{Proceedings of the 44th
  annual international symposium on computer architecture}, 2017, pp. 1--12.

\bibitem{jia2019dissecting}
Z.~Jia, B.~Tillman, M.~Maggioni, and D.~P. Scarpazza, ``Dissecting the
  graphcore ipu architecture via microbenchmarking,'' \emph{arXiv preprint
  arXiv:1912.03413}, 2019.

\bibitem{fick2018analog}
D.~Fick and M.~Henry, ``Analog computation in flash memory for datacenter-scale
  ai inference in a small chip,'' \emph{Hot Chips 2018}, 2018.

\bibitem{gonugondla2020}
S.~K. Gonugondla, C.~Sakr, H.~Dbouk, and N.~R. Shanbhag, ``Fundamental limits
  on the precision of in-memory architectures,'' in \emph{Proceedings of the
  39th International Conference on Computer-Aided Design}, 2020, pp. 1--9.

\bibitem{murmann-survey}
B.~Murmann, ``Adc performance survey 1997–2020.''
  \url{http://web.stanford.edu/~murmann/adcsurvey.html}, 2020, accessed:
  2020-07-12.

\bibitem{garg2021dynamic}
\BIBentryALTinterwordspacing
S.~Garg, J.~Lou, A.~Jain, and M.~Nahmias, ``Dynamic precision analog computing
  for neural networks,'' \emph{arXiv preprint arXiv:2102.06365}, 2021.
  [Online]. Available: \url{https://arxiv.org/abs/2102.06365}
\BIBentrySTDinterwordspacing

\bibitem{oppenheim1970realization}
A.~Oppenheim, ``Realization of digital filters using block-floating-point
  arithmetic,'' \emph{IEEE transactions on audio and electroacoustics},
  vol.~18, no.~2, pp. 130--136, 1970.

\bibitem{song2018computation}
Z.~Song, Z.~Liu, and D.~Wang, ``Computation error analysis of block floating
  point arithmetic oriented convolution neural network accelerator design,'' in
  \emph{Thirty-Second AAAI Conference on Artificial Intelligence}, 2018.

\bibitem{drumond2018training}
M.~Drumond, T.~Lin, M.~Jaggi, and B.~Falsafi, ``Training dnns with hybrid block
  floating point,'' \emph{Advances in Neural Information Processing Systems},
  vol.~31, pp. 453--463, 2018.

\bibitem{mellempudi2017mixed}
N.~Mellempudi, A.~Kundu, D.~Das, D.~Mudigere, and B.~Kaul, ``Mixed
  low-precision deep learning inference using dynamic fixed point,''
  \emph{arXiv preprint arXiv:1701.08978}, 2017.

\bibitem{haowu2020quantization}
\BIBentryALTinterwordspacing
H.~Wu, P.~Judd, X.~Zhang, M.~Isaev, and P.~Micikevicius, ``Integer quantization
  for deep learning inference: Principles and empirical evaluation,''
  \emph{CoRR}, vol. abs/2004.09602, 2020. [Online]. Available:
  \url{https://arxiv.org/abs/2004.09602}
\BIBentrySTDinterwordspacing

\bibitem{jain2019}
\BIBentryALTinterwordspacing
S.~Jain, S.~Venkataramani, V.~Srinivasan, J.~Choi, K.~Gopalakrishnan, and
  L.~Chang, ``Biscaled-dnn: Quantizing long-tailed datastructures with two
  scale factors for deep neural networks,'' in \emph{Proceedings of the 56th
  Annual Design Automation Conference 2019}, ser. DAC '19.\hskip 1em plus 0.5em
  minus 0.4em\relax New York, NY, USA: Association for Computing Machinery,
  2019. [Online]. Available: \url{https://doi.org/10.1145/3316781.3317783}
\BIBentrySTDinterwordspacing

\bibitem{jacob2018quantization}
B.~Jacob, S.~Kligys, B.~Chen, M.~Zhu, M.~Tang, A.~Howard, H.~Adam, and
  D.~Kalenichenko, ``Quantization and training of neural networks for efficient
  integer-arithmetic-only inference,'' in \emph{Proceedings of the IEEE
  conference on computer vision and pattern recognition}, 2018, pp. 2704--2713.

\bibitem{krishnamoorthi2018quantizing}
R.~Krishnamoorthi, ``Quantizing deep convolutional networks for efficient
  inference: A whitepaper,'' \emph{arXiv preprint arXiv:1806.08342}, 2018.

\bibitem{bengio2013estimating}
Y.~Bengio, N.~Léonard, and A.~Courville, ``Estimating or propagating gradients
  through stochastic neurons for conditional computation,'' \emph{arXiv
  preprint arXiv:1308.3432}, 2013.

\bibitem{fan2020training}
A.~Fan, P.~Stock, B.~Graham, E.~Grave, R.~Gribonval, H.~Jégou, and A.~Joulin,
  ``Training with quantization noise for extreme model compression,''
  \emph{arXiv preprint arXiv:2004.07320}, 2020.

\bibitem{baskin2021uniq}
C.~Baskin, N.~Liss, E.~Schwartz, E.~Zheltonozhskii, R.~Giryes, A.~M. Bronstein,
  and A.~Mendelson, ``Uniq: Uniform noise injection for non-uniform
  quantization of neural networks,'' \emph{ACM Transactions on Computer Systems
  (TOCS)}, vol.~37, no. 1--4, pp. 1--15, 2021.

\bibitem{huang2008front}
C.-C. Huang, S.-H. Hung, J.-F. Chung, L.-D. Van, and C.-T. Lin, ``Front-end
  amplifier of low-noise and tunable bw/gain for portable biomedical signal
  acquisition,'' in \emph{2008 IEEE International Symposium on Circuits and
  Systems}.\hskip 1em plus 0.5em minus 0.4em\relax IEEE, 2008, pp. 2717--2720.

\bibitem{chellapilla2006im2col}
K.~Chellapilla, S.~Puri, and P.~Simard, ``High performance convolutional neural
  networks for document processing,'' in \emph{Tenth international workshop on
  frontiers in handwriting recognition}.\hskip 1em plus 0.5em minus 0.4em\relax
  Suvisoft, 2006.

\bibitem{micikevicius2017mixed}
P.~Micikevicius, S.~Narang, J.~Alben, G.~Diamos, E.~Elsen, D.~Garcia,
  B.~Ginsburg, M.~Houston, O.~Kuchaiev, G.~Venkatesh \emph{et~al.}, ``Mixed
  precision training,'' \emph{arXiv preprint arXiv:1710.03740}, 2017.

\bibitem{DBLP:journals/corr/abs-1711-05101}
\BIBentryALTinterwordspacing
I.~Loshchilov and F.~Hutter, ``Fixing weight decay regularization in adam,''
  \emph{CoRR}, vol. abs/1711.05101, 2017. [Online]. Available:
  \url{http://arxiv.org/abs/1711.05101}
\BIBentrySTDinterwordspacing

\bibitem{hooker2021compressed}
\BIBentryALTinterwordspacing
S.~Hooker, A.~Courville, G.~Clark, Y.~Dauphin, and A.~Frome, ``What do
  compressed deep neural networks forget?'' \emph{arXiv preprint
  arXiv:1911.05248}, 2021. [Online]. Available:
  \url{https://arxiv.org/abs/1911.05248}
\BIBentrySTDinterwordspacing

\bibitem{gholami2021quant_survey}
\BIBentryALTinterwordspacing
A.~Gholami, S.~Kim, Z.~Dong, Z.~Yao, M.~W. Mahoney, and K.~Keutzer, ``A survey
  of quantization methods for efficient neural network inference,''
  \emph{CoRR}, vol. abs/2103.13630, 2021. [Online]. Available:
  \url{https://arxiv.org/abs/2103.13630}
\BIBentrySTDinterwordspacing

\bibitem{mckinstry2019percentile}
J.~L. McKinstry, S.~K. Esser, R.~Appuswamy, D.~Bablani, J.~V. Arthur, I.~B.
  Yildiz, and D.~S. Modha, ``Discovering low-precision networks close to
  full-precision networks for efficient inference,'' in \emph{2019 Fifth
  Workshop on Energy Efficient Machine Learning and Cognitive Computing -
  NeurIPS Edition (EMC2-NIPS)}, 2019, pp. 6--9.

\bibitem{welling_2019_ICCV_channelwiseQ}
M.~Nagel, M.~V. Baalen, T.~Blankevoort, and M.~Welling, ``Data-free
  quantization through weight equalization and bias correction,'' in \emph{2019
  IEEE/CVF International Conference on Computer Vision (ICCV)}, 2019, pp.
  1325--1334.

\end{thebibliography}
\bibliographystyle{IEEEtran}

\clearpage
\newpage

\renewcommand{\thepage}{S\arabic{page}} 
\renewcommand{\thesection}{S\arabic{section}}  
\renewcommand{\thetable}{S\arabic{table}}  
\renewcommand{\thefigure}{S\arabic{figure}}
\beginsupplement

\appendix
\section{Analysis of ABFP error and gain in a matrix multiplication}
\label{sec:App_error_analysis}

\begin{figure*}[t]
\centering
\includegraphics[width=\textwidth,height=0.40\textheight]{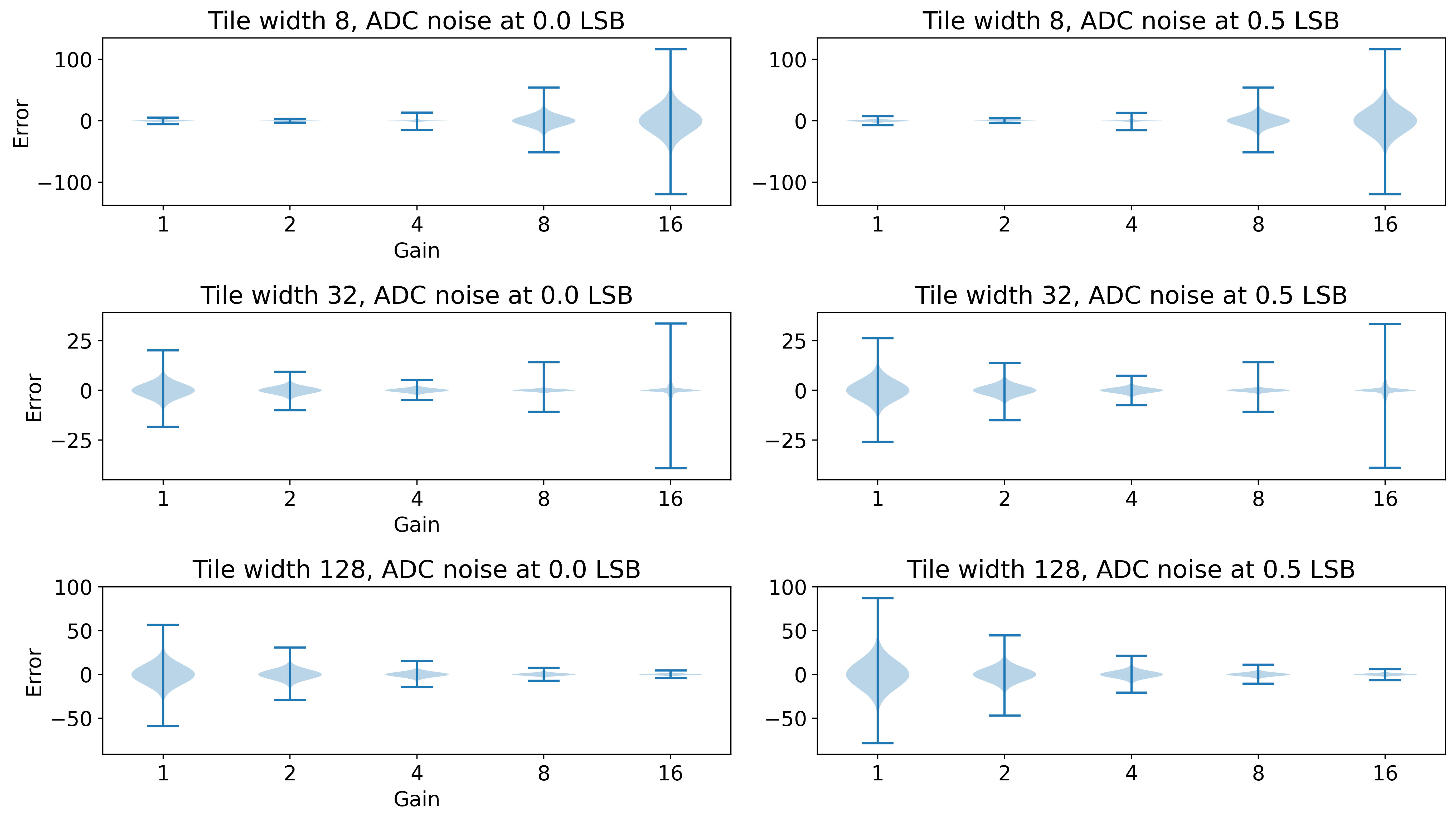}{}
\caption{Effects of gain on error vary with tile size. Smaller tile sizes exhibit less error at lower levels of gain. As gain increases, the magnitudes of the extrema grow, and the variance of the remainder of the errors follow.}
\label{fig:matmul_error_analysis}
\end{figure*}

To characterize the numerical precision of ABFP across parameters of interest, we randomly sample weight and input tensors, multiply them in both \floatfp{} and ABFP, and compute the element-wise difference, $\Delta \mathbf{y}^l$. We do this ten times over tile widths $\{8, 32, 128\}$, gain $\{1, 2, 4, 8, 16\}$,  and ADC noise on LSBs $\{0, 0.5\}$. The bitwidths $b_W/b_X/b_Y$ are fixed at $8/8/8$. The weight matrix of shape $768 \times 768$ and input tensor of shape $16 \times 25 \times 768$ consist of samples drawn from the standard Laplacian and standard normal distributions, respectively. The weight and input shapes are equivalent to those of a BERT Base projection layer with batch size 16 and a sequence length of 25. Our choice of distributions is motivated by commonly-observed distributions of weights and inputs in DNNs.

The distributions of errors are shown in Figure~\ref{fig:matmul_error_analysis}. Error depends on ADC noise, tile size, and gain. As Equation~\eqref{eq:quantization_with_gain_and_noise} suggests, the variance of the error with ADC noise is greater than that without ADC noise. At the smallest tile size, the variance of the error is small and tends to increase with gain; the opposite holds for the largest tile size. This illustrates the utility of gain: it reduces error at larger tile sizes, thereby enabling larger dot products within a single cycle with less information loss. The data show an additional, important dynamic at play. Recall from Section~\ref{sec:overamp} that each doubling of gain drops another more-significant bit and acquires another less-significant bit. The error for a particular matrix-multiplication output in ABFP increases the most when the more-significant bit \textit{dropped} is that output's \textit{most-significant} bit. When this occurs, a large value saturates. Since large values are less frequent in the output of matrix multiplication with operands sampled from a random variable, a step increase in gain initially causes only a small number of values to saturate, at least initially. The rest of the distribution saturates with subsequent increases in gain. This behavior is demonstrated by the extrema of the errors with tile size 32 and gain of 8 and 16.

\section{Supplementary material}

\begin{figure*}[t]
\includegraphics[width=\textwidth]{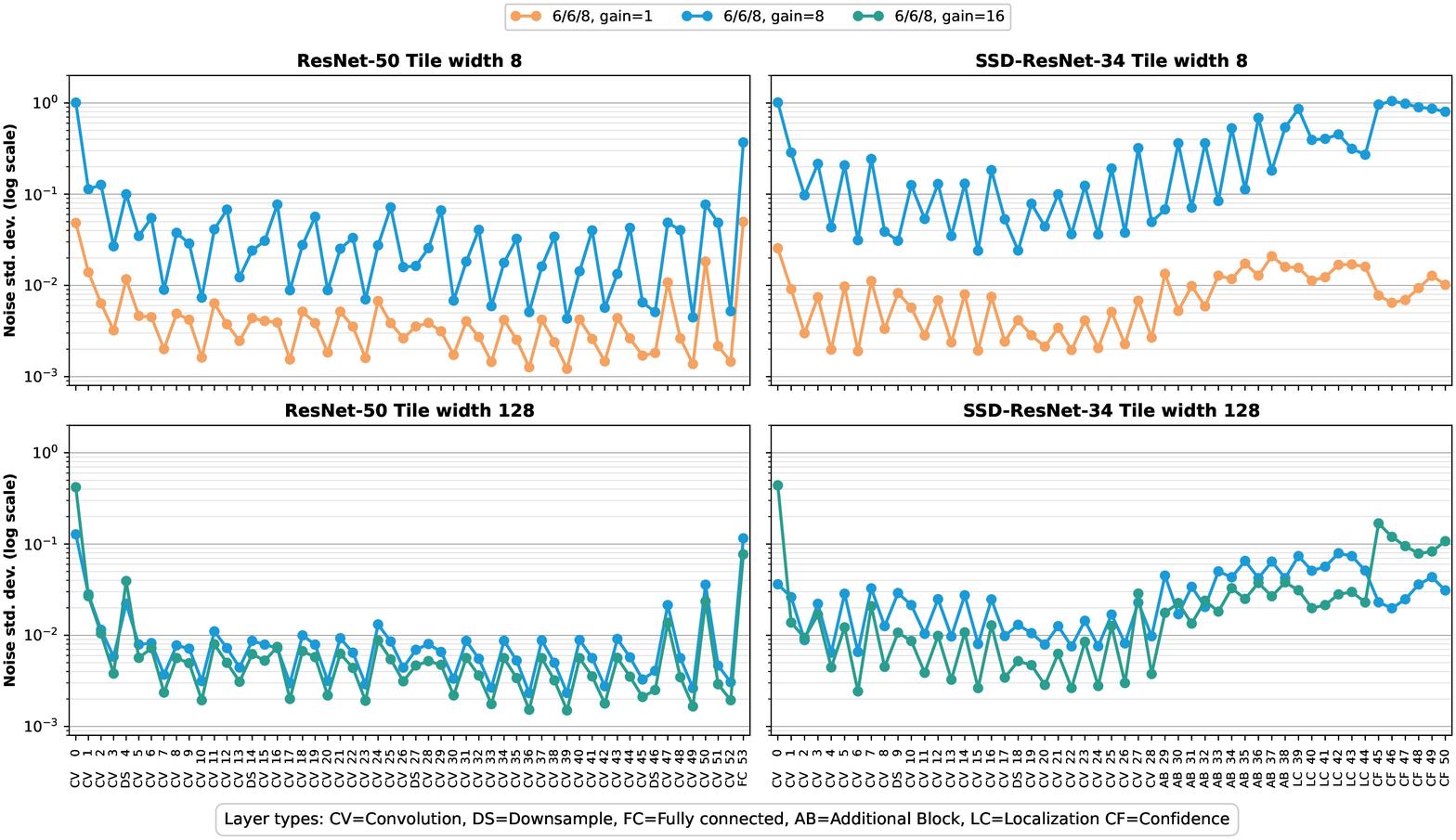}
\vspace{-0.2in}
\caption{Differential noise standard deviations for a subset of layers in ResNet50 and SSD-ResNet34, at tile widths 128 (top) and 8 (bottom) and $b_W/b_X/b_Y = 6/6/8$.}
\label{fig:noise_std_resnet_ssd_app}
\end{figure*}

\begin{itemize}
    \item Table~\ref{tab:ml_perf_links} tabulates the links to the pre-trained Pytorch checkpoints used in Section~\ref{sec:inference_before_finetuning}.
    \item Table~\ref{table:mlperf_std} tabulates the standard deviation of the measured MLPerf\texttrademark{} datacenter inference DNN quality.
    \item Table~\ref{table:retraining_std} tabulates the standard deviation of the top-1 accuracy of the finetuned ResNet50 and mAP score of the SSD-ResNet34 models.
    \item Figure~\ref{fig:noise_std_resnet_ssd_app} plots the differential noise standard deviation for layers in ResNet50 and SSD-ResNet34 for $b_W/b_X/b_Y=6/6/8$.
\end{itemize}

\begin{table*}[h]
    \centering
    \caption{DNNs in the MLPerf\texttrademark{} datacenter inference benchmark and URLs of the checkpoints we used in our experiments. We used the Hugging Face BERT Large checkpoint because its F1 score (93.151\%) on SQuAD v1.1 is higher than that of the checkpoint on the MLPerf\texttrademark{} web site (90.874\%).}
    \vspace{0.1in}
    \begin{tabular}{l p{15.0cm}}
    DNN            & Link to DNN checkpoints \\
    \hline
    ResNet50       & {\small \url{https://doi.org/10.5281/zenodo.4588417}}    \\
    SSD-ResNet34   & {\small \url{https://doi.org/10.5281/zenodo.3236545}}    \\
    3D U-Net       & {\small \url{https://doi.org/10.5281/zenodo.3904106}} \\
    BERT Large     & {\small \url{https://huggingface.co/bert-large-uncased-whole-word-masking-finetuned-squad}} \\
    DLRM           & {\small \url{https://dlrm.s3-us-west-1.amazonaws.com/models/tb00_40M.pt} }     \\
    RNN-T          & {\small \url{https://doi.org/10.5281/zenodo.3662521}} \\
    \end{tabular}
    \label{tab:ml_perf_links}
\end{table*}

\begin{table*}[h!]
\centering
\caption{\label{table:mlperf_std} Standard deviations of MLPerf\texttrademark{} datacenter inference DNN quality metrics for different tile widths, gains and bitwidths. 3D U-Net was run three times due to computational constraints, and the remaining DNNS were run 10 times each.
}
\vspace{0.1in}
\begin{tabular}{r c|c c c c c |c c c c c}

 \multicolumn{2}{c}{} & \multicolumn{5}{c}{$b_W/b_X/b_Y = 6/6/8$} & \multicolumn{5}{c}{$b_W/b_X/b_Y = 8/8/8$} \\ \hline
 \multicolumn{2}{c|}{\bf ResNet50} &  \multicolumn{5}{c|}{\bf Gain} & \multicolumn{5}{c}{\bf Gain} \\
 \multicolumn{2}{c|}{} & 1 & 2 & 4 & 8 & 16 & 1 & 2 & 4 & 8 & 16 \\ \hline
 \multirow{3}{*}{\rotatebox[origin=l]{90}{\parbox{12mm}{\centering \bf Tile \\ width}}} &
    8 & 0.05 & 0.04 & 0.04 & 0.004 & 0.001 & 0.04 & 0.02 & 0.04 & 0.0009 & 0.06 \\
 & 32 & 0.09 & 0.05 & 0.07 & 0.04 & 0.03 & 0.11 & 0.07 & 0.05 & 0.04 & 0.03 \\
 & 128 & 0.02 & 0.14 & 0.08 & 0.08 & 0.06 & 0.04 & 0.09 & 0.06 & 0.05 & 0.03 \\ \hline \hline

 \multicolumn{2}{c|}{\bf SSD-ResNet34} &  \multicolumn{5}{c|}{\bf Gain} & \multicolumn{5}{c}{\bf Gain} \\
 \multicolumn{2}{c|}{} & 1 & 2 & 4 & 8 & 16 & 1 & 2 & 4 & 8 & 16 \\ \hline
 \multirow{3}{*}{\rotatebox[origin=l]{90}{\parbox{12mm}{\centering \bf Tile \\ width}}} &
    8 & 0.02 &  0.02 & 0.01 & 0.009 & 3e-5 & 0.02 & 0.01 & 0.01 & 0.008 & 2.98e-5 \\
 & 32 & 0.04 & 0.04 & 0.03 & 0.02 & 0.01 & 0.05 & 0.04 & 0.03 & 0.01 & 0.01 \\
 & 128 & 0.05 & 0.06 & 0.06 & 0.03 & 0.03 & 0.07 & 0.09 & 0.05 & 0.04 & 0.03 \\ \hline \hline
 
 \multicolumn{2}{c|}{\bf 3D U-Net} &  \multicolumn{5}{c|}{\bf Gain} & \multicolumn{5}{c}{\bf Gain} \\
 \multicolumn{2}{c|}{} & 1 & 2 & 4 & 8 & 16 & 1 & 2 & 4 & 8 & 16 \\ \hline
 \multirow{3}{*}{\rotatebox[origin=l]{90}{\parbox{12mm}{\centering \bf Tile \\ width}}} &
    8 & 0.0085 & 0.0100 & 0.0017 & 0.0035 & 0.0035 & 0.0023 & 0.0065 & 0.0368 & 0.1311 & 0.0000 \\
 & 32 & 0.0115 & 0.0098 & 0.0225 & 0.0040 & 0.0017 & 0.0377 & 0.0150 & 0.0101 & 0.0050 & 0.0023 \\
 & 128 & 0.0682 & 0.0493 & 0.0051 & 0.0000 & 0.0040 & 0.0270 & 0.2455 & 0.0234 & 0.0106 & 0.0058 \\ \hline \hline

 \multicolumn{2}{c|}{\bf RNN-T} &  \multicolumn{5}{c|}{\bf Gain} & \multicolumn{5}{c}{\bf Gain} \\
 \multicolumn{2}{c|}{} & 1 & 2 & 4 & 8 & 16 & 1 & 2 & 4 & 8 & 16 \\ \hline
 \multirow{3}{*}{\rotatebox[origin=l]{90}{\parbox{12mm}{\centering \bf Tile \\ width}}} &
    8 & 0.004 &  0.005 & 0.04 & 0.02 & 0.00 & 0.02 & 0.01 & 0.003 & 0.02 & 0.00 \\
 & 32 & 0.02 & 0.03 & 0.001 & 0.01 & 0.01 & 0.03 & 0.07 & 0.03 & 0.02 & 0.03 \\
 & 128 & 0.03 & 0.06 & 0.03 & 0.01 & 0.01 & 0.12 & 0.09 & 0.13 & 0.01 & 0.06 \\ \hline \hline
 
 \multicolumn{2}{c|}{\bf BERT-Large} &  \multicolumn{5}{c|}{\bf Gain} & \multicolumn{5}{c}{\bf Gain} \\
 \multicolumn{2}{c|}{} & 1 & 2 & 4 & 8 & 16 & 1 & 2 & 4 & 8 & 16 \\ \hline
 \multirow{3}{*}{\rotatebox[origin=l]{90}{\parbox{12mm}{\centering \bf Tile \\ width}}} &
    8 & 0.0589 &  0.0415 & 0.0314 & 0.0396 & 0.0446 & 0.0303 & 0.0458 & 0.0319 & 0.0604 & 0.0324 \\
 & 32 & 0.1299 & 0.0772 & 0.0853 & 0.0726 & 0.1337 & 0.1413 & 0.0893 & 0.0799 & 0.0456 & 0.1038 \\
 & 128 & 0.0945 & 0.1787 & 0.1645 & 0.0555 & 0.0876 & 0.1580 & 0.1383 & 0.1422 & 0.1300 & 0.0618 \\ \hline \hline
 
 \multicolumn{2}{c|}{\bf DLRM} &  \multicolumn{5}{c|}{\bf Gain} & \multicolumn{5}{c}{\bf Gain} \\
 \multicolumn{2}{c|}{} & 1 & 2 & 4 & 8 & 16 & 1 & 2 & 4 & 8 & 16 \\ \hline
 \multirow{3}{*}{\rotatebox[origin=l]{90}{\parbox{12mm}{\centering \bf Tile \\ width}}} &
    8 & 0.0005 &  0.0000 & 0.0000 & 0.0000 & 0.0000 & 0.0005 & 0.0000 & 0.0000 & 0.0000 & 0.0000 \\
 & 32 & 0.0005 & 0.0000 & 0.0000 & 0.0000 & 0.0000 & 0.0005 & 0.0000 & 0.0005 & 0.0000 & 0.0000 \\
 & 128 & 0.0000 & 0.0000 & 0.0000 & 0.0000 & 0.0000 & 0.0000 & 0.0000 & 0.0000 & 0.0000 & 0.0000 \\ \hline
 
\end{tabular}
\end{table*}

\begin{table*}[h!]
\centering
\caption{\label{table:retraining_std} Standard deviations for finetuning results in Table \ref{table:retraining}.}
\vspace{0.1in}
\begin{tabular}{ccc}
 & \multicolumn{2}{c}{$b_W/b_X/b_Y$} \\ \cline{2-3}
Finetuning technique & $6/6/8$ & $8/8/8$ \\ \hline
\multicolumn{3}{l}{\bf ResNet50} \\
\multicolumn{1}{l}{QAT} &  0.071 & 0.043 \\
\multicolumn{1}{l}{DNF} & 0.065  & 0.035  \\ \hline
\multicolumn{3}{l}{\bf SSD-ResNet34} \\
\multicolumn{1}{l}{QAT} & 0.034  & 0.037  \\
\multicolumn{1}{l}{DNF} & 0.026  & 0.034  \\ \hline
\end{tabular}
\end{table*}

\end{document}